\documentclass[letterpaper, 10 pt, conference]{ieeeconf}  % Comment this line out if you need a4paper

\IEEEoverridecommandlockouts                              % This command is only needed if 
\title{\LARGE \bf
LiSnowNet: Real-time Snow Removal\\for LiDAR Point Clouds
}

\author{
    Ming-Yuan Yu$^{1}$, Ram Vasudevan$^{2}$, and Matthew Johnson-Roberson$^{3}$% <-this % stops a space
    \thanks{This work was supported by a grant from Ford Motor Company via the Ford-UM Alliance under Award N022884. (Corresponding author: Ming-Yuan Yu.)}
    \thanks{$^{1}$M.-Y Yu is with Robotics Institute, University of Michigan, Ann Arbor, MI 48109, USA
        {\tt\footnotesize myyu@umich.edu}}%
    \thanks{$^{2} $R. Vasudevan is with the Department of Mechanical Engineering, the University of Michigan, Ann Arbor, MI 48109, USA
        {\tt\footnotesize ramv@umich.edu}}%
    \thanks{$^{3} $M. Johnson-Roberson is with the Department of Naval Architecture and Marine Engineering at the University of Michigan, Ann Arbor, MI 48109, USA
        {\tt\footnotesize mattjr@umich.edu}}%
}

\usepackage{times}

\usepackage[bookmarks=true]{hyperref}

\usepackage{booktabs}
\usepackage{amsfonts}
\usepackage{amsmath}
\usepackage{subfig}
\usepackage{graphicx}
\usepackage{perl_acronyms}

\usepackage[
    style=ieee,
    doi=false,
    isbn=false,
    url=false,
    eprint=false,
    backend=bibtex,
    natbib=true
    ]{biblatex}
\begin{document}

\maketitle
\thispagestyle{empty}
\pagestyle{empty}

\begin{abstract}
\acp{LiDAR} have been widely adopted to modern self-driving vehicles, providing 3D information of the scene and surrounding objects. However, adverser weather conditions still pose significant challenges to LiDARs since point clouds captured during snowfall can easily be corrupted. The resulting noisy point clouds degrade downstream tasks such as mapping. Existing works in de-noising point clouds corrupted by snow are based on nearest-neighbor search, and thus do not scale well with modern \acp{LiDAR} which usually capture $100k$ or more points at $10$Hz. In this paper, we introduce an unsupervised de-noising algorithm, LiSnowNet, running $52\times$ faster than the state-of-the-art methods while achieving superior performance in de-noising. Unlike previous methods, the proposed algorithm is based on a deep convolutional neural network and can be easily deployed to hardware accelerators such as GPUs. In addition, we demonstrate how to use the proposed method for mapping even with corrupted point clouds.
\end{abstract}

\IEEEpeerreviewmaketitle

\section{Introduction}
\label{sec:introduction}

\acp{LiDAR} are classified as active sensors, emitting pulsed light waves and utilizing the returning pulse to calculate the distances between it and the surrounding objects. This makes \acp{LiDAR} suitable during both daytime and nighttime, providing detailed 3D measurements of the surrounding scenes. Such measurements can be easily found in modern datasets~\cite{Geiger2013IJRR,Sun2020CVPR,John2020CORR,fong2021panoptic} as frame-by-frame point clouds, commonly sampled at $10$Hz, and has been used for 3D object detection~\cite{lang2019pointpillars,shi2020points}, semantic segmentation~\cite{cheng2021s3net,cortinhal2020salsanext}, and mapping~\cite{legoloam2018}. 

\begin{figure}[t]
    \centering
    \subfloat[CADC~\cite{pitropov2021canadian}]{% trim={left, bottom, right, top}
        \includegraphics[width=0.47\linewidth,trim={25mm 120mm 30mm 48mm},clip]{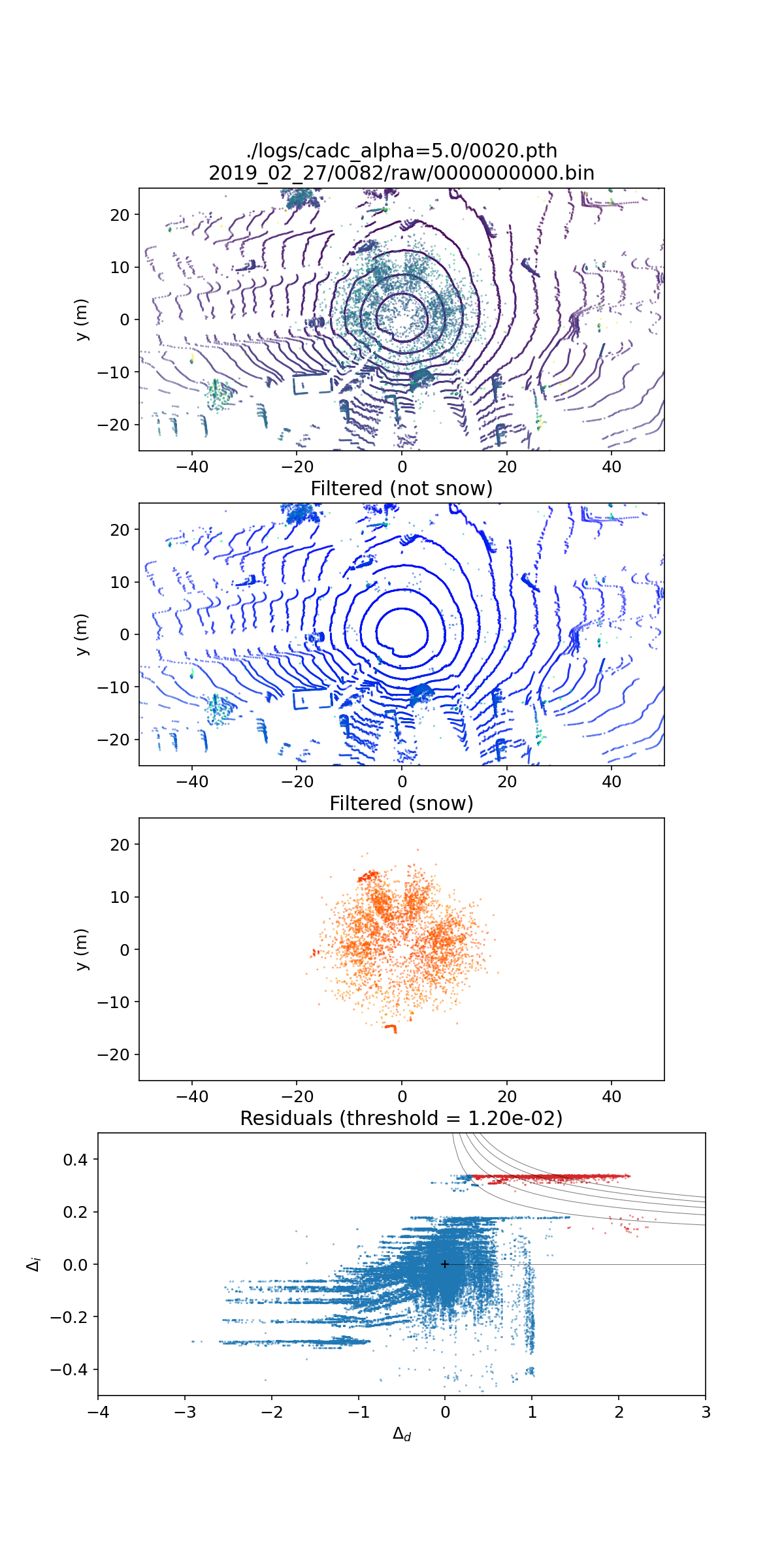}
        \label{fig:denoised_cadc}
    }
    \subfloat[WADS~\cite{bos2020autonomy}]{
        \includegraphics[width=0.47\linewidth,trim={25mm 120mm 30mm 48mm},clip]{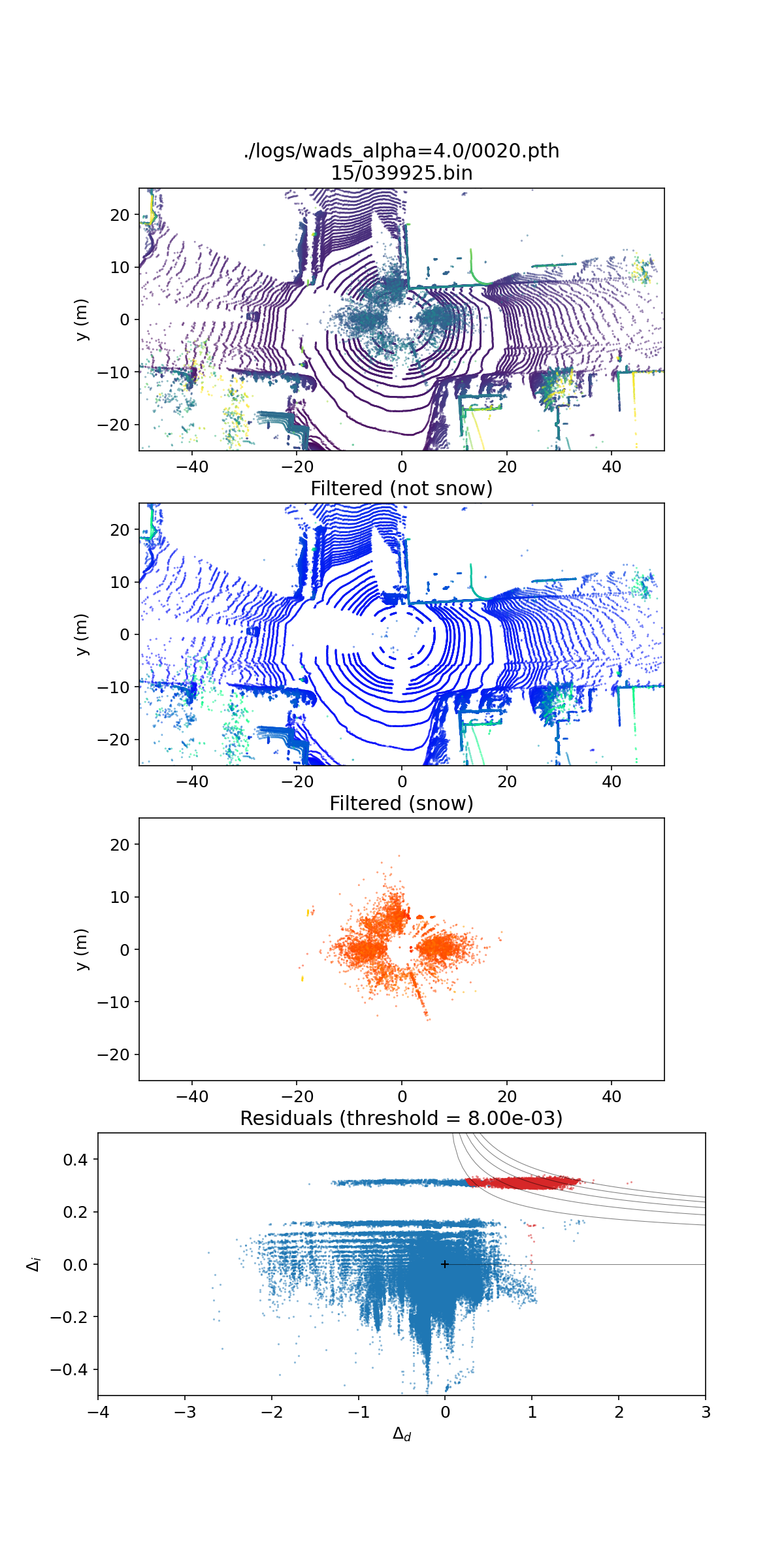}
        \label{fig:denoised_wads}
    }
    \caption{Point clouds corrupted by snowflakes from two datasets~\cite{pitropov2021canadian,bos2020autonomy}, and the de-noised results using the proposed method. From top to bottom: Raw point clouds, de-noised point clouds (predicted as non-snow), and the ghost measurements. Both axes are in meters.}
    \label{fig:denoised}
\end{figure}

Though \acp{LiDAR} provide more accurate 3D measurements compared to radar, the measurements are prone to degrade under adverse weather conditions. Unlike radars, which see through fog, snowflakes and rain droplets, \acp{LiDAR} are greatly affected by particles in the air~\cite{kutila2016automotive,charron2018dror,kurup2021dsor}. Specifically during snowfall, the pulsed light waves hit the snowflakes and return to the sensor as \textit{ghost} measurements, as shown in Fig. \ref{fig:denoised_cadc} and \ref{fig:denoised_wads}. It is crucial to remove such measurements and unveil the underlying geometry of the scene for applications like mapping.

Recently, there is an increasing number of large-scale datasets containing adverse weather conditions. The most relevant ones are the \ac{CADC} Dataset~\cite{pitropov2021canadian} and the \ac{WADS}~\cite{bos2020autonomy}. Denoising algorithms~\cite{charron2018dror,kurup2021dsor} have also been developed alongside with them. To the best of our knowledge, all of the unsupervised de-noising algorithms are based on nearest neighbor search and have difficulty operating in real-time even with a moderate $64-$beam \ac{LiDAR} (about $100k$ points/frame). Therefore, it is critical to introduce an alternative formulation which does not rely on nearest neighbor search.
 
An alternative way to represent point clouds is using \textit{range images}~\cite{Heinzler2020PointCloudDeNoising,cortinhal2020salsanext}, as shown in Fig. \ref{fig:ri_distance} and \ref{fig:ri_intensity}. Each range image has two channels, and the pixel values are 1) the Euclidean distances in the body coordinate and 2) the intensities. With this representation, we can leverage common algorithms in image processing, including the \ac{FFT}, the \ac{DWT}, and the \ac{DoG}~\cite{marr1980theory}. \acp{CNN} with pooling, batch normalization~\cite{DBLP:journals/corr/IoffeS15} and dropout~\cite{pmlr-v48-gal16} can also be applied without any modification. More importantly, these operations can perform very efficiently on modern hardware such as GPUs.
 
However, developing a \ac{CNN} for de-noising point clouds poses its own challenges. The most obvious one is that obtaining point-wise labels to isolate snowflakes from the scene is usually prohibitively expensive on large-scale datasets. Thus, the formulation needs to be either unsupervised or self-supervised to exploit unlabeled data. With this in mind, we propose a novel network, the \textit{LiSnowNet}, which can be trained only with unlabeled point clouds. 

We assume the noise-free range images are sparse under some transformation, and design loss functions that relies purely on quantifying the sparsity of the data. The $L_1$ norm is used as a proxy to sparsity, and encourages sparse coefficients during training. In addition, the \ac{DWT} and the \ac{IDWT} are used for down sampling and up scaling within the network instead of pooling and convolution transpose, respectively. The proposed network is evaluated on a subset of \ac{WADS}~\cite{bos2020autonomy}, showing similar or superior performance in de-noising compared to other state-of-the-art methods while running $52$ times faster.

This paper is organized as follows: Section \ref{sec:related_work} lists related work in datasets with adverse weather conditions, existing de-noising methods, and backgrounds in compressed sensing with sparse representations. Section \ref{sec:method} goes through the formulation of the proposed network and loss functions in detail, including how we train the network in an unsupervised fashion. Section \ref{sec:evaluation} introduces the metrics for comparing the performance of the proposed network and other methods. Section \ref{sec:results} quantitatively shows the results of the aforementioned metrics, as well as qualitatively presents the resulting map from an off-the-shelf mapping algorithm. Lastly, Section \ref{sec:conclusions} summarizes the contributions and advantages of the proposed method.
\section{Related Work}
\label{sec:related_work}

\subsection{Datasets}

Large-scale datasets for autonomous driving~\cite{Geiger2013IJRR,fong2021panoptic,John2020CORR,Sun2020CVPR} become more ubiquitous nowadays. Many of the recent ones contain \ac{LiDAR} data collected during adverse weather conditions, targeting challenging scenarios like fog, rain, and snow.

\citet{Heinzler2020PointCloudDeNoising} collected a dataset with points clouds in a weather-controlled climate chamber, simulating rain and fog in front of a vehicle. The dataset provides point-wise labels with three classes: clear, rain, and fog, where the ``clear'' points are the ones which are not caused by the adverse weather.

\citet{pitropov2021canadian} published the \ac{CADC} dataset which contains $32754$ point clouds under various snow precipitation levels. The point clouds were collected from a top-mounted HDL-32e and were grouped into sequences where each sequence has about $100$ consecutive frames. Though the \ac{CADC} dataset has 3D bounding boxes for vehicles and pedestrians, we do not use them in our work as the proposed method is designed to be unsupervised. 

Recently \citet{bos2020autonomy} released the \ac{WADS} with only $1828$ point clouds which is significantly smaller than \ac{CADC}. But unlike \ac{CADC}, each point cloud in \ac{WADS} has point-wise labels. There are $22$ classes, including ``active falling snow'' and ``accumulated snow''.  

Since the target application of the proposed network is snow removal, only \ac{CADC} and \ac{WADS} are used for training and evaluation.

\begin{figure*}[ht]
    \centering
    \subfloat[The distance channel.]{% trim={left, bottom, right, top}
        \includegraphics[width=0.98\linewidth]{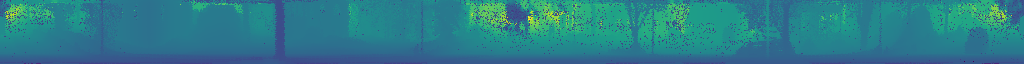}
        \label{fig:ri_distance}
    }
    
    \subfloat[The intensity channel.]{% trim={left, bottom, right, top}
        \includegraphics[width=0.98\linewidth]{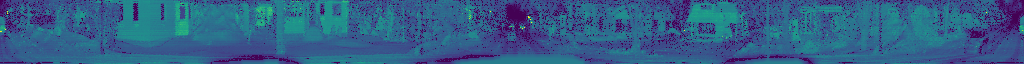}
        \label{fig:ri_intensity}
    }
    
    \subfloat[The log-magnitude of the \ac{FFT} coefficients of the distance channel.]{% trim={left, bottom, right, top}
        \includegraphics[width=0.98\linewidth]{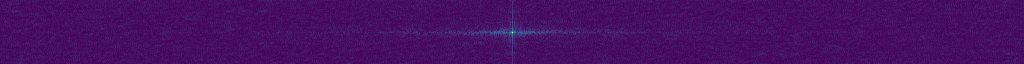}
        \label{fig:ri_distance_fft}
    }
    
    \subfloat[The magnitude of the \ac{DWT} coefficients of the distance channel.]{% trim={left, bottom, right, top}
        \includegraphics[width=0.98\linewidth]{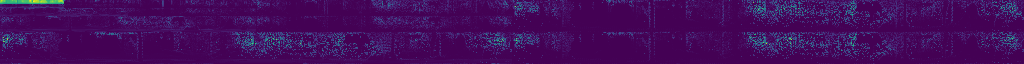}
        \label{fig:ri_distance_dwt}
    }
    
    \caption{\textbf{(a) \& (b)}: a range image $\tilde{I}_k$ from \ac{WADS}~\cite{bos2020autonomy}. \textbf{(c) \& (d)}: the corresponding sparse coefficients of the distance channel.}
    \label{fig:ri_example}
\end{figure*}

\subsection{De-noising Point Clouds Corrupted by Snow}

Similar to the proposed method, the WeatherNet~\cite{Heinzler2020PointCloudDeNoising} projects point clouds onto a spherical coordinate as range images, which we go into detail in Section \ref{sec:method:preprocessing}. With this representation and the dataset released alongside with the WeatherNet, they formulated the de-noising problem as supervised multi-class pixel-wise classification. The network is constructed by a series of LiLaBlock~\cite{piewak2018boosting} and is trained with point-wise class labels. The WeatherNet is able to remove rain and fog, but it is unclear how it performs on snow. Additionally, training the WeatherNet requires point-wise label. This prevents the WeatherNet from taking advantage of large-scale unlabeled point cloud datasets such as \ac{CADC}.

The 2D median filter is successful in removing salt and pepper noise. Once we convert point clouds into range images, we can apply the median filter to them. As shown in Fig. \ref{fig:ri_example}, the pixels corresponding to snow are relatively darker than their nearby pixels in both channels. They are darker in the distance channel because the background is further away from the ego vehicle; in addition, snowflakes tend to absorb the energy of the beams which results in dark pixels in the intensity channel. An asymmetrical median filter can be applied to remove just the ``peppers'' (i.e. snow, isolated dark pixels) while leaving other pixels untouched.

The \ac{DROR}~\cite{charron2018dror} filter removes the noise by first constructing a $k-d$ tree, and count the number of neighbors of each point using dynamic search radii. The idea is that the density of the points from the scene is inversely proportional to the distance, while the density of the points from snowflakes follows a different distribution. The search radius of each point is dynamically adjusted by its distance to the \ac{LiDAR} center accordingly. If a point does not have enough number of neighbors within its search radius, it is classified as an outlier (i.e. snow).

The \ac{DSOR}\cite{kurup2021dsor} filter also relies on nearest neighbor search, but it calculates the mean and variance of relative distances given a fixed number of neighbors of each point. In other words, it estimates the local density around each point, and outliers can then be filtered out. \ac{DSOR} is shown to be slightly faster than \ac{DROR} while having a similar performance in de-noising.

Both \ac{DROR} and \ac{DSOR} require nearest neighbor search which prevents them from being real-time. A moderate \ac{LiDAR} like the HDL‐64E generates around $100k$ points at $10$Hz. Querying this many points with a $k-d$ tree of this size takes hundreds of milliseconds on modern hardware. On the other hand, WeatherNet~\cite{Heinzler2020PointCloudDeNoising} and our method work with image-like data, which will be discussed later in Section \ref{sec:method:preprocessing}. Common tools in image processing, such as \ac{FFT}, \ac{DWT}, and \acp{CNN}, can be applied directly and very efficiently on GPUs. This allows us to develop a real-time de-noising algorithm that runs orders of magnitude faster than \ac{DROR} and \ac{DSOR}.

We compare our method with \ac{DROR}, \ac{DSOR}, and a $5\time 5$ median filter in Section \ref{sec:results}, since their implementations are publicly available.

\subsection{Sparse Representations}
\label{sec:representations}

Let $u, v$ be two signals in $\mathbb{R}^n$. The signal $u$ is said to be \textit{sparser than} $v$ under a transformation $\mathcal{T}$ if $\mathcal{T}(u)$ has more zero coefficients than $\mathcal{T}(v)$. In other words, 
\begin{equation}
    \|\mathcal{T}(u)\|_0 < \|\mathcal{T}(v)\|_0,
    \label{eq:l0-sparsity}
\end{equation}
\noindent
where $\|\cdot\|_0$ is the $L_0$ norm.

Though the $L_0$ norm directly quantifies the sparsity of a signal, using it in practice is problematic because it is not convex. Instead, the $L_1$ norm is widely used as a proxy to quantify sparsity and recover corrupted signals~\cite{candes2008enhancing}. Thus we relax Eq. \ref{eq:l0-sparsity} and get
\begin{equation}
    u \mbox{ is sparser than } v \Leftrightarrow \|\mathcal{T}(u)\|_1 < \|\mathcal{T}(v)\|_1
    \label{eq:l1-sparsity}
\end{equation}
\noindent
under some transformation $\mathcal{T}$, where $\|\cdot\|_1$ is the $L_1$ norm.

The basic assumption of our work is that real-world \textit{clean} signals are sparse under some transformation $\mathcal{T}$. Both \ac{DWT} and \ac{FFT} are known to generate sparse representations on real-world multi-dimensional grid data like audio and images, meaning that the Fourier and Wavelet coefficients of natural signals are mostly very close zero.

Typically, a signal corrupted by additive noise becomes less sparse in \ac{FFT} and \ac{DWT} than the underlying clean one. This is due to the fact that common noises, such as isolated spikes or Gaussian, have wide bandwidths in the frequency domain, as shown in Fig. \ref{fig:ri_distance_fft} and \ref{fig:ri_distance_dwt}. Therefore de-noising is possible if the underlying clean signal is sufficiently sparse in Fourier or Wavelet. The de-noising can be formulated as maximizing the sparsity of the transformed signal. 

Assuming that a noisy signal $v$ is roughly equally sparse under $m$ different transformations $\mathcal{T}_1, \ldots, \mathcal{T}_m$, we aim to solve
\begin{equation}
    \min_u \lambda\|u - v\|_q + \frac{1}{m} \sum_{j=1}^{m}\|\mathcal{T}_j(u)\|_1,
    \label{eq:sparse_l1}
\end{equation}
\noindent
where $u$ is the underlying clean signal, $\|\cdot\|_q$ is the $L_q$ norm, and $\lambda\in\mathbb{R}$ is the weight keeping $u$ close to $v$. The value $q$ is commonly set to be either $1$ or $2$, depending on the structure of the noise: $1$ if the noise itself is sparse, and $2$ otherwise.

However, solving Eq. \ref{eq:sparse_l1} directly is computationally intensive for any reasonably sized signal. Thus we propose to use a \ac{CNN} to approximate the solution and use Eq. \ref{eq:sparse_l1} as the groundwork of our loss functions. More details about our formulation can be found in Section \ref{sec:method:network} and \ref{sec:method:loss}. 

\begin{figure*}[ht]
    \centering
    \subfloat[Architecture of LiSnowNet]{% trim={left, bottom, right, top}
        \includegraphics[width=0.6\linewidth]{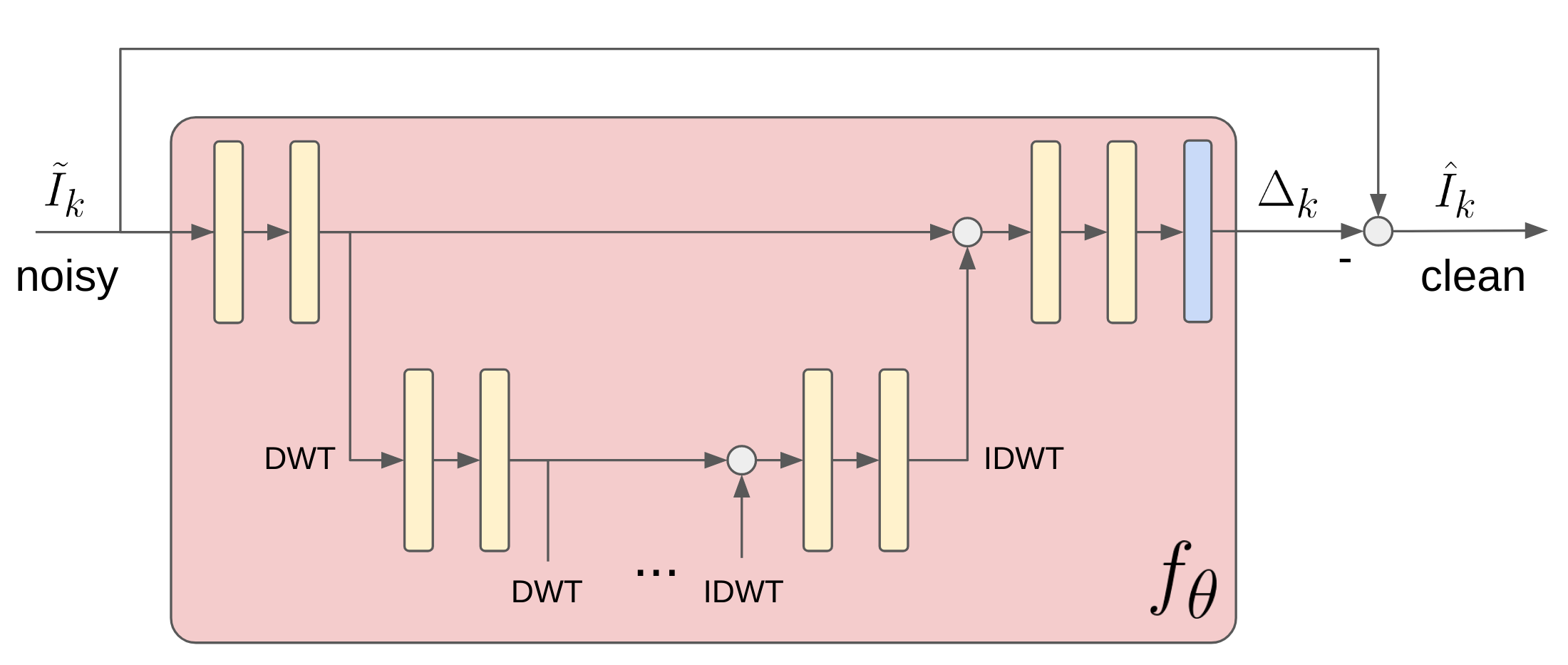}
        \label{fig:network_arch}
    }
    \subfloat[Modified Residual Block]{
        \includegraphics[width=0.3\linewidth]{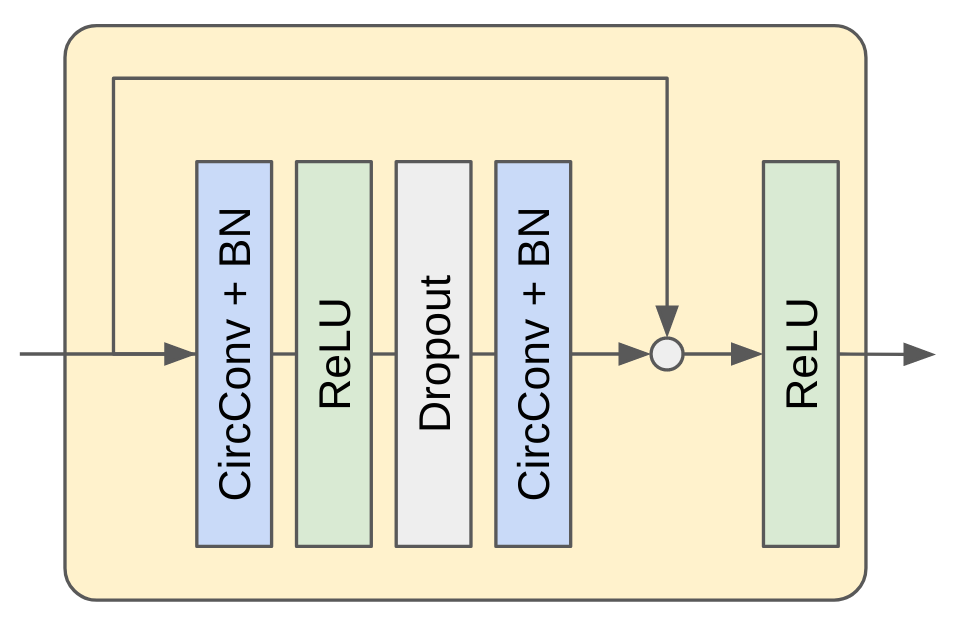}
        \label{fig:resblock}
    }
    \caption{The proposed network, LiSnowNet $f_{\theta}$. The top level takes a range image $\tilde{I}_k$ with size $h\times w\times 2$ as an input. Each level consists multiple modified residual blocks (yellow) with skip connections. Similar to the MWCNN~\cite{DBLP:journals/corr/abs-1907-03128}, the down sampling and up scaling operations are \ac{DWT} and \ac{IDWT}, respectively. The output $\Delta_k$ is the residual image representing the noise component of $\tilde{I}_k$. Subtracting $\Delta_k$ from $\tilde{I}_k$ results in the desired clean image $\hat{I}_k$.}
    \label{fig:network_arch_full}
\end{figure*}

\section{Method}
\label{sec:method}

\subsection{Preprocessing}
\label{sec:method:preprocessing}

The first step is to convert point clouds into range images. Given a point $p = \begin{bmatrix}x & y & z \end{bmatrix}^\top \in \mathbb{R}^3$ in the $k-$th point cloud in the dataset and its corresponding intensity value $i\in[0,1]$, we calculate the following values:
\begin{equation}
    \begin{split}
        d &:= \|p\|_2 \in\mathbb{R}_+, \\
        \phi &:= \sin^{-1}\left(\frac{z}{d}\right) \in \left[-\frac{\pi}{2}, \frac{\pi}{2}\right], \\
        \psi &:= \tan^{-1}\left(\frac{y}{x}\right) \in \left[-\pi, \pi\right),
    \end{split}
    \label{eq:euclidean2spherical}
\end{equation}
\noindent
where $d$ is the distance from the \ac{LiDAR} center, $\phi$ is the inclination angle, and $\psi$ is the azimuth angle. By discretizing the inclination and azimuth angles within the \ac{FOV} of the \ac{LiDAR}, we project each point within a point cloud onto a spherical coordinate, resulting a range image $I_k\in\mathbb{R}^{h\times w \times 2}_+$, where $k\in\mathbb{Z}$ is the frame index, $h$ is the vertical resolution, and $w$ is the horizontal resolution. Throughout this paper $h$ is set to be the number of beams of the \ac{LiDAR} and $w$ is set to be a constant $2048$. The first channel of the range image $I_k$ is the distance $d$ of each point, and the second channel is the corresponding intensity value $i$.

The second step is to squash the range images to a proper scale. A key observation is that the ghost measurements mainly concentrate within a distance of $25$ meters, as shown in Fig. \ref{fig:denoised}, whereas the maximum distance of a \ac{LiDAR} easily exceeds $150$ meters. We scale the range images $I_k$ with an element-wise function $g_1:\mathbb{R}^{h\times w\times 2}_+ \mapsto \mathbb{R}^{h\times w\times 2}_+$, defined as
\begin{equation}
    \label{eq:distortion}
    g_1(I_k) := \sqrt[3]{I_k},
\end{equation}
\noindent
to amplify the relative importance of points near the ego vehicle while preserving its ordering in distances. The function $g_1$ also enhances the contrast between the noise and the scene in the intensity channel, since the intensity values of the snowflakes are almost always $0$ whereas the scene usually has positive intensity values.

Finally, not every pixel has a value due to the lack of points at certain directions like the sky and some transparent surfaces. The void pixels is processed with a series of operations, collectively denoted as $g_2:\mathbb{R}^{h\times w\times 2}_+ \mapsto \mathbb{R}^{h\times w\times 2}_+$, before entering the network. In sequence, the operations are:

\begin{enumerate}
    \item $3\times3$ dilation to fill isolated void pixels.
    \item Filling the remaining void pixels with the value $(\mu_k + \sigma_k)$, where $\mu_k\in\mathbb{R}^{h\times2}_+$ and $\sigma_k\in\mathbb{R}^{h\times2}_+$ are the row-wise mean and standard deviation, respectively.
    \item Subtraction with a $3\times3$ \ac{DoG} to reduce the amplitude of isolated noisy pixels.
    \item $7\times7$ average pooling to smooth out the image.
\end{enumerate}

Note that $g_2$ does not modify any pixel with a valid measurement. Only the void pixels are altered. An example of the a resulting image
\begin{equation}
    \tilde{I}_k := (g_2\circ g_1)(I_k)   
\end{equation}
\noindent
is shown in Fig. \ref{fig:ri_distance} and \ref{fig:ri_intensity}.

\subsection{Network Architecture}
\label{sec:method:network}

The proposed network, LiSnowNet $f_\theta:\mathbb{R}^{h\times w\times 2}_+ \mapsto \mathbb{R}^{h\times w\times 2}$, is based on the MWCNN~\cite{DBLP:journals/corr/abs-1907-03128} with several key modifications. First, all convolution layers are replaced by residual blocks~\cite{DBLP:journals/corr/HeZRS15} with two circular convolution layers. In addition, a dropout layer is placed after the first ReLU activation in each residual block to further regularize the network. Lastly, the number of channels is dramatically reduced -- the first level only has $8$ channels compared to the $160$ channels in MWCNN. The number of channels of the higher levels are also reduced proportionally. These modifications are critical for processing sparse representations (i.e. \ac{FFT} and \ac{DWT} coefficients) of panoramic range images in real-time.

The proposed network is designed to produce the \textit{residual image} $\Delta_k := f_\theta\left(\tilde{I}_k\right)$, satisfying
\begin{equation}
    \hat{I}_k := \tilde{I}_k - \Delta_k,
    \label{eq:residual}
\end{equation}

\noindent
where $\hat{I}_k$ is the de-noised range image representing the geometry of the underlying scene.

\subsection{Loss Functions}
\label{sec:method:loss}

Let $\mathcal{F}:\mathbb{R}^{h\times w\times 2} \mapsto \mathbb{R}^{h\times w\times 2}$ be the real-valued \ac{FFT} and $\mathcal{W}:\mathbb{R}^{h\times w\times 2} \mapsto \mathbb{R}^{h\times w\times 2}$ be the \ac{DWT} with the Haar basis. With the formulation in Eq. \ref{eq:sparse_l1} and \ref{eq:residual}, we designed three novel loss functions, two of which quantify the sparsity of the range image, while the third one quantifies the sparsity of the noise.
\begin{equation}
    \begin{split}
        \mathcal{L}_\mathcal{F} &:= \frac{1}{N} \sum_{k=1}^N \left\|\log\left(\left|\mathcal{F}\left(\hat{I}_k\right)\right| + 1\right)\right\|_1, \\
        \mathcal{L}_\mathcal{W} &:= \frac{1}{N} \sum_{k=1}^N \left\|\mathcal{W}\left(\hat{I}_k\right)\right\|_1, \\
        \mathcal{L}_\Delta &:= \frac{1}{N} \sum_{k=1}^N \left\| \Delta_k \right\|_1, \\
    \end{split}
    \label{eq:losses}
\end{equation}

\noindent
where $N$ is the number of point clouds in the training set, and $|\cdot|$ is the element-wise absolute value.

Note that $\mathcal{L}_\mathcal{F}$ uses the log-magnitude of the Fourier coefficients. Since $\log(1 + \epsilon) \approx \epsilon$ for some small $\epsilon$, it prevents the network from focusing too much on the low-frequency part of the spectrum while pertaining the properties of the $L_1$ norm. The total loss function $\mathcal{L}$ can thus be constructed as
\begin{equation}
    \mathcal{L} := \alpha\cdot\frac{\mathcal{L}_\mathcal{F} + \mathcal{L}_\mathcal{W}}{2} + (1 - \alpha)\cdot \mathcal{L}_\Delta,
    \label{eq:total_loss}
\end{equation}

\noindent
where $\alpha\in(0,1)$ is a hyperparameter balancing the sparsity of the range image and the residual.

\subsection{Training}
\label{sec:method:training}

The network is trained on \ac{CADC} and \ac{WADS} with each dataset split into $80\%$ for training and $20\%$ for validation, where the minimum unit is a sequence. The batch size is $32$ and $8$ on the two datasets, respectively. Each training session runs $20$ epochs with the Adam optimizer and an initial learning rate of $0.001$. The learning rate is updated at the end of each epoch with a learning decay of $0.89$.

\subsection{Prediction}
\label{sec:method:prediction}

After the network is trained, the residual images $\Delta_k~\forall k$ are used for filtering. Let $\Delta^d_k\in\mathbb{R}^{h\times w}$ be the distance channel and $\Delta^i_k\in\mathbb{R}^{h\times w}$ be the intensity channel of the residual image $\Delta_k$ at time $k$. We define the decision boundary primarily in the residual space. A point $p$ with residuals $\delta^d_k\in\Delta^d_k$ and $\delta^i_k\in\Delta^i_k$ is classified as snow (i.e. positive) if it satisfies all the following conditions:

\begin{itemize}
    \item Being the foreground (i.e. $p$ is closer than nearby pixels). $$\delta^d_k > 0$$
    \item Absorbing most of the energy of \ac{LiDAR} beams (i.e. $p$ is darker than nearby pixels). $$\delta^i_k > 0$$
    \item The residuals of $p$ are not sparse. $$\left(\delta^d_k\right)^{n_d} \cdot \left(\delta^i_k\right)^{n_i} > \bar{\delta}$$
\end{itemize}

\noindent
where $n_d\in\mathbb{R}_+$ and $n_i\in\mathbb{R}_+$ are the shape parameters of the primary decision boundary, and $\bar\delta\in\mathbb{R}_+$ is a threshold value.
\begin{table}
    \centering
    \begin{tabular}{@{}lrrrr@{}}
        \toprule
        & Precision & Recall & IoU & Runtime [ms] \\ \midrule
        \ac{DROR}~\cite{charron2018dror} & 0.8533 & 0.9508 & 0.8163 & 1071.2 \\
        \ac{DSOR}~\cite{kurup2021dsor} & 0.5311 & \textbf{0.9811} & 0.5242 & 352.5 \\
        Median Filter ($5\times5$) & 0.8306 & 0.8082 & 0.6946 & 7.7 \\
        LiSnowNet-$L_2$ (ours) & 0.8716 & 0.9597 & 0.8415 & \textbf{6.8} \\
        LiSnowNet-$L_1$ (ours) & \textbf{0.9249} & 0.9501 & \textbf{0.8812} & \textbf{6.8} \\
        \bottomrule
    \end{tabular}
    \caption{De-noising results.}
    \label{tbl:results}
\end{table}

\section{Evaluation}
\label{sec:evaluation}

Quantitatively, the proposed network and other two state-of-the-art methods~\cite{charron2018dror,kurup2021dsor} are evaluated with three common metrics for binary classification on a subset of \ac{WADS}~\cite{bos2020autonomy}, since it is the only publicly available dataset containing point-wise labels with snow at the moment of writing this paper. Let $TP$ be the number of true positive points, $FP$ be the number of false positive points, and $FN$ be the number of false negative points. The metrics are as follows:

\begin{itemize}
    \item Precision: $\frac{TP}{TP + FP}$
    \item Recall: $\frac{TP}{TP + FN}$
    \item \ac{IoU}: $\frac{TP}{TP + FP + FN}$
\end{itemize}

We also record the average runtime on a modern desktop computer with a Ryzen 2700X and an RTX 3060.

In addition, qualitative results in mapping with sequences of de-noised point clouds are presented. We use the \ac{RTK} poses from \ac{CADC} and \ac{LeGO-LOAM} to demonstrate the effectiveness of building maps from de-noised point clouds. Both \ac{RTK} and \ac{LeGO-LOAM} are able to reconstruct the environment from the raw and de-noised point clouds, but the difference in the quality of the resulting map is significant. 
\section{Results}
\label{sec:results}

Compared to the baselines, the proposed network (labeled as LiSnowNet-$L_1$) yields similar or superior performance in de-noising, as shown in TABLE \ref{tbl:results}. The recall is $0.0310$ lower than \ac{DSOR}, while the precision and \ac{IoU} are noticeably higher than all other methods. 

More importantly, our network is $52\times$ faster than \ac{DSOR} and $158\times$ faster than \ac{DROR} with about $100k$ points per frame. Considering that the sampling rate of \acp{LiDAR} are usually at $10$Hz, the proposed method is the best one that can operate in real-time.

\begin{figure*}
    \centering
    \subfloat{% trim={left, bottom, right, top}
        \includegraphics[width=0.95\linewidth,trim={0mm 9mm 0mm 0mm},clip]{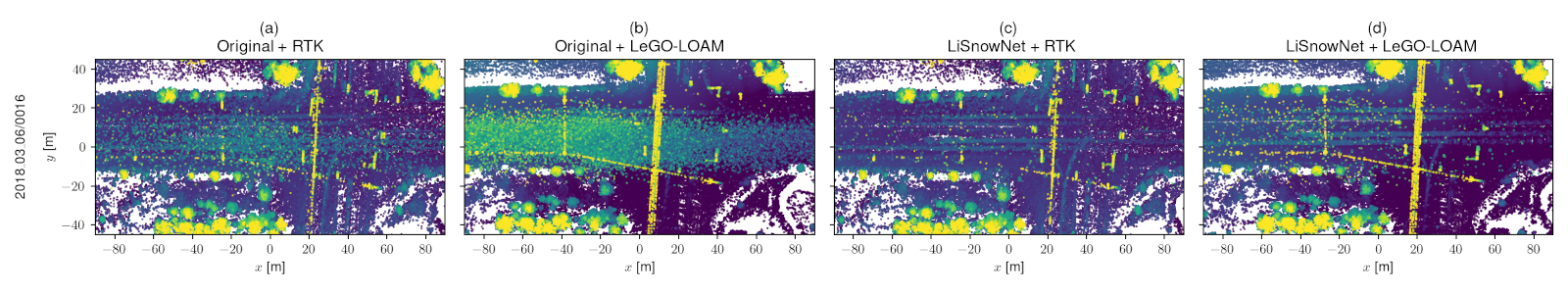}
    }
    
    \subfloat{% trim={left, bottom, right, top}
        \includegraphics[width=0.95\linewidth,trim={0mm 9mm 0mm 10mm},clip]{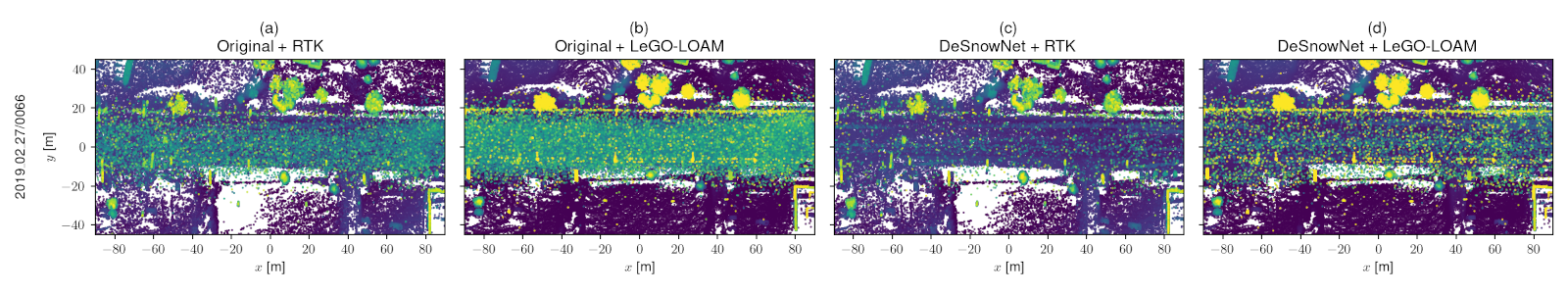}
    }
    
    \subfloat{% trim={left, bottom, right, top}
        \includegraphics[width=0.95\linewidth,trim={0mm 9mm 0mm 10mm},clip]{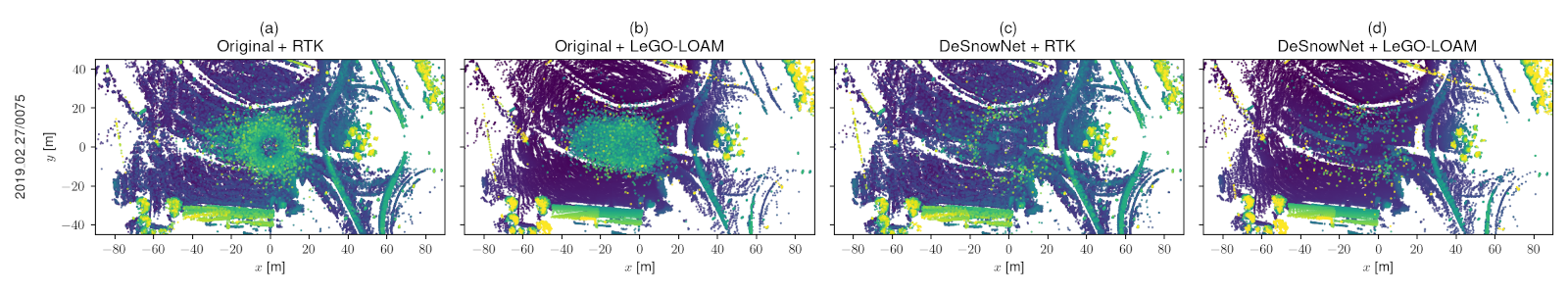}
    }
    
    \subfloat{% trim={left, bottom, right, top}
        \includegraphics[width=0.95\linewidth,trim={0mm 9mm 0mm 10mm},clip]{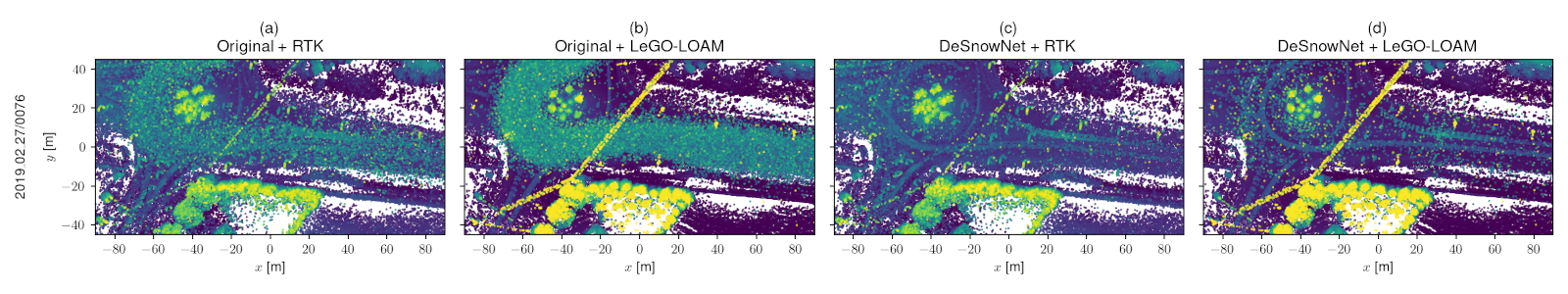}
    }
    
    \subfloat{% trim={left, bottom, right, top}
        \includegraphics[width=0.95\linewidth,trim={0mm 9mm 0mm 10mm},clip]{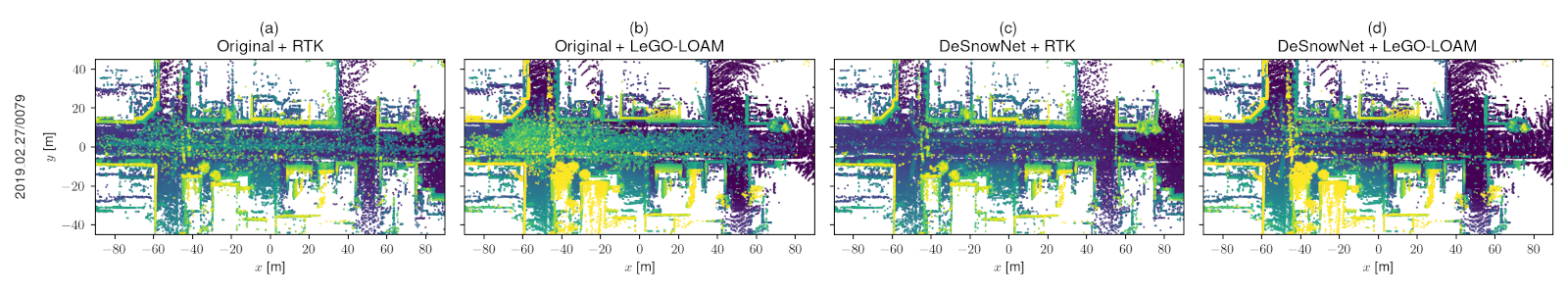}
    }
    
    \subfloat{% trim={left, bottom, right, top}
        \includegraphics[width=0.95\linewidth,trim={0mm 0mm 0mm 10mm},clip]{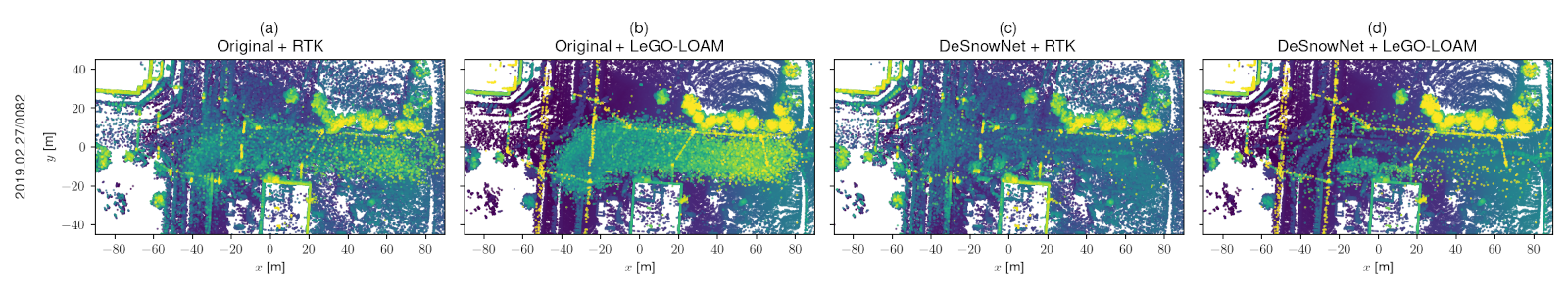}
    }
    \caption{Bird's-eye view of the reconstructed maps in \ac{CADC}~\cite{pitropov2021canadian}. \textbf{(a) \& (b)}: from raw point clouds; \textbf{(c) \& (d)} from de-noised point clouds. Each row corresponds to a sequence in \ac{CADC}. The point clouds processed by the proposed method are able to generate much cleaner maps using the \ac{RTK} poses as well as using \ac{LeGO-LOAM}~\cite{legoloam2018}. The colors indicate the height (along the $z$ axis) of the points.}
    \label{fig:lego-loam}
\end{figure*}

The de-noised point clouds also enable building clean maps using either poses from the \ac{RTK} system or \ac{LeGO-LOAM}, as shown in Fig. \ref{fig:lego-loam}. \ac{LeGO-LOAM} is able to generate near identical trajectories with both the raw and de-noised point clouds, but the map with the former shows a large number of ghost points above where the ego vehicle traveled. By contrast, resulting map with de-noise point clouds is much cleaner, containing a minimal amount of noise and faithfully reflecting the occupancy of the scene.

\section{Ablation Study}
\label{sec:ablation}

\begin{table}
    \centering
    \begin{tabular}{@{}ccrrr@{}}
        \toprule
        Optimal $\alpha$ & $\beta$ & Precision & Recall & \ac{IoU} \\ \midrule
        0.9779 & 0.10 & 0.9363 & 0.9417 & 0.8840 \\
        0.9558 & 0.30 & 0.9378 & 0.9338 & 0.8783 \\
        0.9779 & 0.50 & 0.9249 & \textbf{0.9501} & 0.8812 \\
        0.9688 & 0.70 & \textbf{0.9457} & 0.9250 & 0.8768 \\
        0.9779 & 0.90 & 0.9444 & 0.9432 & \textbf{0.8928} \\
        \bottomrule
    \end{tabular}
    \caption{Sensitivity with respect to the weights of sparse transformations.}
    \label{tbl:fft_vs_dwt}
\end{table}

\subsection{\texorpdfstring{$L_1$ versus $L_2$}{L1 versus L2}}

As stated in Section \ref{sec:representations}, the $L_1$ norm widely is used as a proxy to sparsity of real-world signals. To further justify the usage of $L_1$ norm as oppose to some other common norms, such as the $L_2$ norm, we train the same network but replace the $L_1$ norm with the $L_2$ norm in the loss functions. We label the trained networks LiSnowNet-$L_1$ and LiSnowNet-$L_2$ to reflect the norm used during training, as shown in TABLE~\ref{tbl:results}.

We see that LiSnowNet-$L_2$ already performs better than \ac{DROR} in all metrics. Using the $L_1$ norm brings further improvements in precision and \ac{IoU} while keeping a similar recall. The results indicate that using the $L_1$ norm as a proxy to sparsity is more effective than using the $L_2$ norm.

\subsection{FFT versus DWT}

Both \ac{FFT} and \ac{DWT} are linear operators which are know to transform natural images to sparse representations in the frequency domain. The key difference is that \ac{FFT} captures global features across the whole image, while \ac{DWT} captures local features at various scales of the image.

We investigate the sensitivity of the contributions of individual transformation by varying the ratio of $\mathcal{L}_\mathcal{F}$ and $\mathcal{L}_\mathcal{W}$ in Eq. \ref{eq:total_loss}. The modified total loss $\mathcal{L}$ thus becomes
\begin{equation}
    \mathcal{L} := \alpha \cdot (\beta \cdot \mathcal{L}_\mathcal{F} + (1 - \beta)\cdot \mathcal{L}_\mathcal{W}) + (1 - \alpha)\cdot \mathcal{L}_\Delta,
\end{equation}
\noindent
where $\beta\in[0,1]$ determines the relative contributions of \ac{FFT} and \ac{DWT}. We train the LiSnowNet using this loss function with five different $\beta$ values ranging from 0.1 to 0.9, as shown in TABLE \ref{tbl:fft_vs_dwt}. The optimal $\alpha$ is obtained by grid search for each $\beta$ value.

We see that the variations of the performance remain reasonably small, with the maximum difference being $0.0251 = 0.9501 - 0.9250$ in recall. This number is way smaller than the variation between different methods presented in TABLE \ref{tbl:results}, indicating that the proposed network is robust to the selection of sparse transformations if $\alpha$ and $\beta$ are properly tuned.
\section{Conclusions}
\label{sec:conclusions}

We propose a deep \ac{CNN}, LiSnowNet, for de-noising point clouds corrupted under adverse weather conditions. The proposed network can be trained without any labeled data, and generalizes well to multiple datasets. The network is able to process $100k$ points under $7$ms while yielding superior performance in de-noising compared to the state-of-the-art method. It also enhances the quality of down stream tasks such as mapping during snowfall.

\printbibliography

\end{document}